\documentclass[10pt]{article}
\usepackage[a4paper, total={6.2in, 9.7in}]{geometry}
\pdfoutput=1

\usepackage{times}
\usepackage{soul}
\usepackage{url}
\usepackage[hidelinks]{hyperref}
\usepackage[utf8]{inputenc}
\usepackage[small]{caption}
\usepackage{graphicx}
\usepackage{amsmath}
\usepackage{amsthm}
\usepackage{booktabs}
\usepackage{algorithm}
\usepackage{algorithmic}
\usepackage[switch]{lineno}
\usepackage[disable]{todonotes}
\usepackage{svg}
\usepackage{tikz}
\usetikzlibrary{positioning}
\usetikzlibrary{shapes}
\usepackage{subcaption}
\usepackage{authblk}

\title{ Ontology Pre-training for Poison Prediction
% ...
}

\author[1]{Martin Glauer}
\author[1]{Fabian Neuhaus}
\author[1]{Till Mossakowski}
\author[2,3]{Janna Hastings}
\affil[1]{Institute for Intelligent Cooperating Systems, Otto-von-G\"uricke University Magdeburg}
\affil[2]{Faculty of Medicine, Institute for Implementation Science in Health Care, University of Zurich}
\affil[3]{School of Medicine, University of St. Gallen}

\begin{document}

\maketitle

\begin{abstract}  
    Integrating human knowledge into neural networks has the potential to improve their robustness and interpretability. We have developed a novel approach to integrate knowledge from ontologies into the structure of a Transformer network which we call ontology pre-training: we train the network to predict membership in ontology classes as a way to embed the structure of the ontology into the network, and subsequently fine-tune the network for the particular prediction task. We apply this approach to a case study in predicting the potential toxicity of a small molecule based on its molecular structure, a challenging task for machine learning in life sciences chemistry. Our approach improves on the state of the art, and moreover has several additional benefits. First, we are able to show that the model learns to focus attention on more meaningful chemical groups when making predictions with ontology pre-training than without, paving a path towards greater robustness and interpretability. Second, the training time is reduced after ontology pre-training, indicating that the model is better placed to learn what matters for toxicity prediction with the ontology pre-training than without. This strategy has general applicability as a neuro-symbolic approach to embed meaningful semantics into neural networks. 
\end{abstract}


\section{Introduction}

Deep neural networks have recently led to breakthrough performance for a wide range of tasks such as protein folding \cite{jumper_highly_2021} and image generation \cite{rombach_high-resolution_2022}. 
However, they still suffer from challenges in generalisability, robustness, and interpretability. Approaches that incorporate human knowledge alongside learning from data, which have been called hybrid, 
knowledge-aware or \textit{informed} \cite{von_rueden_informed_2021}, 
have the potential to improve the correspondence between what the model 
learns and the structure of the human world, which in turn allows the 
model to learn more generalisable representations from smaller datasets. 

Human knowledge is carefully curated into ontologies \cite{dessimoz_primer_2017,neuhaus_ontology_2022}, 
making them a prime candidate source of knowledge to 
incorporate into learning. Many different approaches 
have already been developed with the objective of 
harnessing prior knowledge to improve machine learning. The 
most common approach is enrichment of the training 
data with additional information from ontologies (see Section~\ref{sec:relatedWork}). 
In this paper we present a novel methodology, which 
uses an ontology, namely the Chemical Entities of 
Biological Interest (ChEBI), to create a pre-training task for a Transformer model (see Section~\ref{sec:methods}). 
This pre-training task consists of predicting 
superclasses in ChEBI's taxonomic 
hierarchy for molecule classes represented by input chemical structures. Thus, during this pre-training the model 
learns to recognise categories of chemical entities 
that are chemically meaningful. After the ontology 
pre-training the model is fine-tuned for the task of toxicity prediction 
using the dataset from the well-known Tox21 challenge \cite{huang_tox21challenge_2016}. This dataset consists of 12 different toxicity endpoints, including 7 nuclear receptor signals and 5 stress response indicators. 

As we show in Section~\ref{sec:results}, for the purpose of toxicity prediction the ontological pre-training step showed the following benefits: 
First, the model converges faster during fine-tuning. 
Second, an inspection of the attention heads indicates that the model pays attention to chemical structures that correspond to structural chemical annotations that are associated with classes in ChEBI. Since ChEBI classes represent chemical categories that are meaningful to humans, this connection improves the interpretability of the model's predictions.  
Third, the predictive performance is improved significantly compared to 
the performance without pre-training. Indeed, our ontology pre-trained model outperforms the state of the art for toxicity prediction on the Tox21 dataset from structures without additional input features (see Section~\ref{sec:relatedWork}).  

These results seem to indicate that the ontological pre-training is enabling the model to learn some of the knowledge that is represented by the ontology. 
However, there are important limitations with respect to the knowledge that is learned by the model. Further, our proposed methodology is only applicable to ontologies that either contain rich structural annotations or which are associated with suitable datasets that link the input datatype intended for the learning task to the ontology classes. We will discuss these limitations in  Section~\ref{ssec:limitations}.

\section{Methods}
\label{sec:methods}
The usual process used to train a transformer-based model consists of two steps: pre-training and fine-tuning. The intention behind the pre-training step is to give the model a solid foundation training in the kind of data that will be used in downstream tasks in order for it to gain a preliminary understanding of the target domain that can the transferred to more specific tasks later. Most transformer-based architectures are built for language problems, and the respective pre-training is often limited to masked-language tasks (BERT \cite{DBLP:conf/naacl/DevlinCLT19}, RoBERTa) or token discrimination tasks (Electra). This kind of training enables the model to learn the syntactic relations between words, but it does not get any information about their semantics, aside from context similarity when words are interchangeably used in similar contexts.

\begin{figure}
\centering
\scalebox{0.9}{
\begin{tikzpicture}[]
    \tikzstyle{process} = [draw,rounded corners=8pt,align=center]
    \tikzstyle{model} = [chamfered rectangle, chamfered rectangle xsep=2cm, draw ,align=center]
    \node[draw] (chebi) at (5,1) {data:ChEBI};
    \node[draw] (pubchem) at (0,0) {data:Unlabeled molecules};
    \node[draw] (tox21) at (2.5,5) {data:Tox21};
    \node[process] (pre)  at (0,1) {Train: Electra Pre-training};
    \node[model] (premo)  at (0,2.5) {Pre-trained Model};
    \node[process] (ontpre)   at (5,2.5) {Train: Ontology-based Task};
    \node[model] (ontpremo)  at (5,4) {Ontology \\ Pre-trained Model};
    \node[process] (toxpre_sem)   at (5,6) {Toxicity Prediction};
    \node[process] (toxpre_nosem)   at (0,6) {Toxicity Prediction};

    \draw[->] (pubchem) -- (pre);
    \draw[->] (pre) -- (premo);
    \draw[->] (premo) -- (ontpre);
    \draw[->] (chebi) -- (ontpre);
    \draw[->] (ontpre) -- (ontpremo);
    \draw[->] (ontpremo) -- (toxpre_sem);
    \draw[->] (premo) -- (toxpre_nosem);
    \draw[->] (tox21) -- (toxpre_nosem);
    \draw[->] (tox21) -- (toxpre_sem);
    
\end{tikzpicture}}
\caption{Training stack for standard training and ontology pre-training}
\label{fig:stack}
\end{figure}
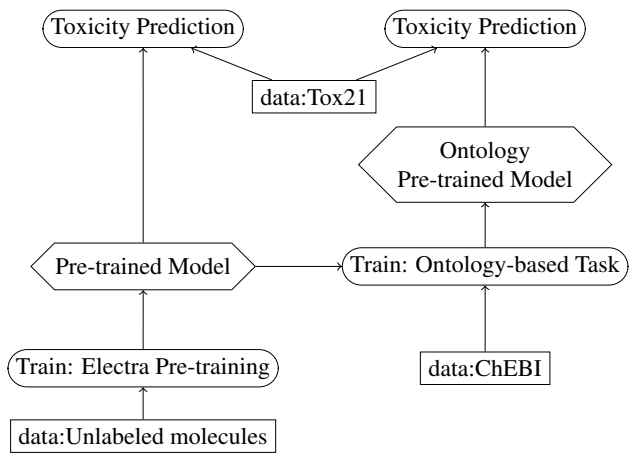

\begin{table}
\centering
\begin{tabular}{l|l}
\textbf{Hyperparameter} & \textbf{Value}\\ \hline
  Vocab. size & 1,400 \\
  Hidden size & 256 \\
  Num. of attention heads & 8 \\
  Num. of hidden layers & 6 \\
  Epochs & 100 \\
  Learning Rate & $1e^{-4}$ \\
  Optimizer & Adamax \\
\end{tabular}
\caption{Hyperparameters shared by all models.}
\label{tbl:Hyperparameters}
\end{table}

In this study, we introduced an additional ontology pre-training stage after the usual pre-training and before the fine-tuning. We present a case-study for the use of ontology pre-training to improve chemical toxicity prediction. Figure~\ref{fig:stack} depicts the process using the boxology notation \cite{van2021modular} for the novel approach as well as a comparison approach without ontology pre-training which serves as our baseline. In the remainder of this section, we will detail the setup for each individual step. All models are based on the Electra model as implemented in the ChEBai\footnote{\url{https://github.com/MGlauer/ChEBai}} tool used in previous work \cite{glauer_interpretable_2022}. All models share the hyperparameters detailed in Table~\ref{tbl:Hyperparameters} with different classification heads for different tasks.

\subsection{Step 1: Standard Pre-Training}
\label{ssec:pre-training}

The first step of this architecture is based on the pre-training mechanism for Electra \cite{clark2020electra}. This process extends the general masked-language model (MLM) used to pre-train transformers by a competing discriminator that aims to predict which tokens have been replaced as part of the MLM task. For the molecular data that serves as the input for the prediction task, the language of the molecular representations is SMILES \cite{weininger_smiles_1988}, a molecular line notation in which atoms and bonds are represented in a linear sequence, and the masking of tokens affects sub-elements within the molecule. The generator part of Electra uses a simple linear layer with as softmax to predict the token that was most likely masked, while the discriminator uses the same setup, but one additional dense layer to guess which token had been replaced by the generator.

In our case-study, we use the same dataset for pre-training that has been used in a previous task on ontology extension \cite{glauer_interpretable_2022}. This dataset consists of 365,512 SMILES representations for molecular structures that have been extracted from PubChem \cite{sayers_pubchem_2005} and the ChEBI ontology \cite{hastings_chebi_2016}. Notably,  152.205 of these substances are known to be hazardous substances as they are associated with a hazard class annotation in PubChem. 

\subsection{Step 2: Ontology Pre-Training}

The standard pre-training teaches the network the foundations of how chemical structures are composed and represented in SMILES strings, but it does not give the network any insights into which parts of these molecules may be important and chemically active. These functional properties are semantic information that are used by experts to distinguish classes of molecules within the ChEBI ontology. They are therefore inherently encoded in the subsumption relations of the ontology. We used the subsumption hierarchy to create a dataset for an ontology-based classification task by extracting all subclasses of `molecular entity' that had at least 100 subclasses with SMILES strings attached to them. This resulted in a collection of 856 classes. We then collected all classes in ChEBI that had a SMILES string attached to them and annotated them with their subsumption relation for each of the 856 classes. The resulting dataset is similar to the ones used in \cite{Hastings2021} and \cite{glauer_interpretable_2022}, but covers a wider range of classes as labels (856 instead of 500) and also a larger number of molecules (129,187 instead of 31,280). We then use the pre-trained model from Step 1 to predict the superclasses of a class of molecules based on its annotated SMILES string.

\subsection{Step 3: Toxicity prediction}

In order to assess the impact of this additional ontology-based pre-training step, we compare the model that resulted from the ontology pre-training in step 2 with the one from step 1 that did not receive that training. This comparison is based on each model's performance on toxicity prediction using the Tox21 dataset. This dataset was created by the National Center for Advanced Translational Sciences (NCATS) of the National Institutes of Health (NIH), and constitutes a widely used benchmark for research on machine learning for toxicity prediction from small molecules \cite{huang_tox21challenge_2016}. Notably, there are currently two versions of the Tox21 dataset available in benchmarks. The first one originates from the Tox21 Challenge that was conducted in 2014. This dataset consists of three different subsets, one for training and two for different steps of the challenge evaluation. In our study, we will use the ``testing dataset" that was used for an initial ranking of models as a validation set and the dataset that was used for the final evaluation as our test set. This version of the Tox21 dataset suffers from several issues regarding the consistency and quality of different entries \cite{idakwo2020structure}. A more modern version of this dataset has been published as part of the MoleculeNet benchmark \cite{wu2018moleculenet}. This version of Tox21 consists of only of 7,831 molecules. We split this dataset into a training (85\%), validation (7.5\%) and test set (7.5\%). There are, however, still two major challenges that need to be considered when working with this dataset. First, the number of datapoints is rather low. Molecular structures are complex graphs, which makes it hard for any model to derive a sufficient amount of information from this dataset alone. Second, the information available in the dataset is not complete: a substantial amount of toxicity information is missing in this dataset. There are 4,752 molecules for which at least one of the 12 kinds of toxicity is not known. In the prior literature, this issue has been approached in different ways. Some approaches perform data cleaning which limits the number of available data points even further, e.g. \cite{idakwo2020structure} excluded all datapoints that contained any missing labels. We decided to keep these datapoints as part of our dataset, but exclude the missing labels from the calculation of all loss functions and metrics. Any outputs that the model generates for these missing labels is considered as correct and does not influence the training gradient. This approach allows the network to fill these gaps autonomously.

Preliminary results showed that both models were prone to overfitting, when used with the same settings as the model from step 2. We found that this behaviour could be prevented by using strong regularisations. The final experiments used the Adamax optimizer with dropouts on input embeddings of 0.2, dropouts on hidden states of 0.4 and L2-regularisation of 0.0001.
All data and code used in this study is publicly available.\footnote{\url{https://doi.org/10.5281/zenodo.7548313}}

\section{Results}
\label{sec:results}
\subsection{Predictive performance}

\begin{figure*}
\captionsetup[subfigure]{aboveskip=-5pt,belowskip=-2pt}
    \begin{subfigure}{0.48\textwidth}
            
    \includegraphics[width=0.89\textwidth]{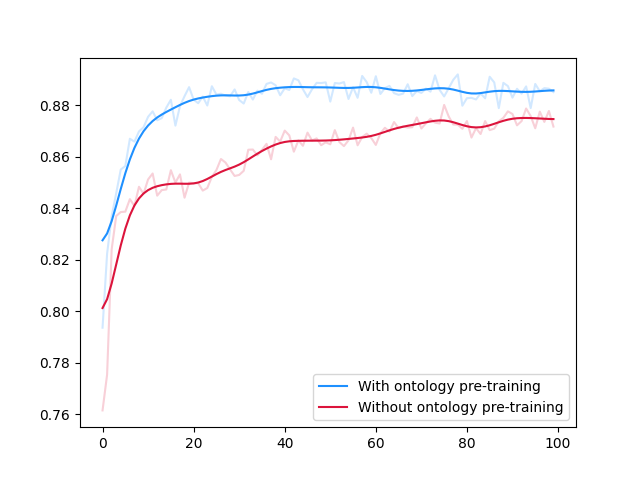}
    \caption{ ROC-AUC / MoleculeNet Tox21 dataset }
    \label{fig:ar-moleculenet}
    \end{subfigure}
    \hfill
    \begin{subfigure}{0.48\textwidth}
    \includegraphics[width=0.89\textwidth]{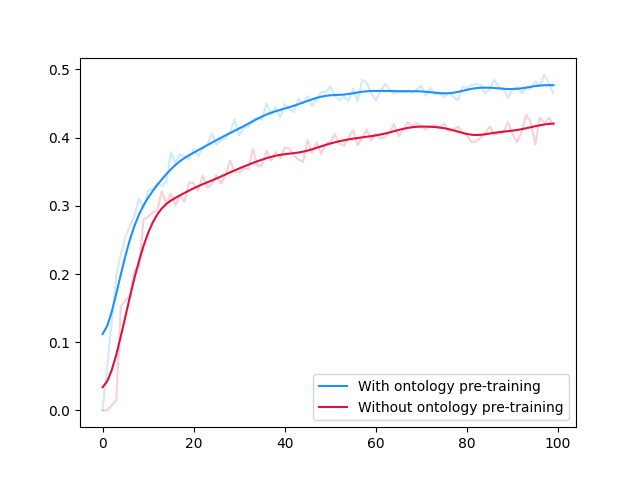}
    \caption{F1 score (micro) / MoleculeNet Tox21 dataset}
    \label{fig:f1-moleculenet}
    \end{subfigure}
    
    \noindent\vspace{-0.5em}
    \begin{subfigure}{0.48\textwidth}
    \includegraphics[width=0.89\textwidth]{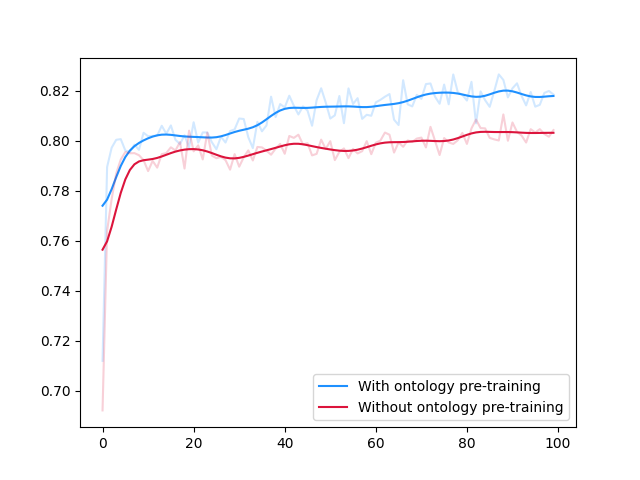}

    \caption{ROC-AUC / original TOX21 dataset }
    \label{fig:ar-challenge}
    \end{subfigure}
    \hfill
     \begin{subfigure}{0.48\textwidth}
    \includegraphics[width=0.89\textwidth]{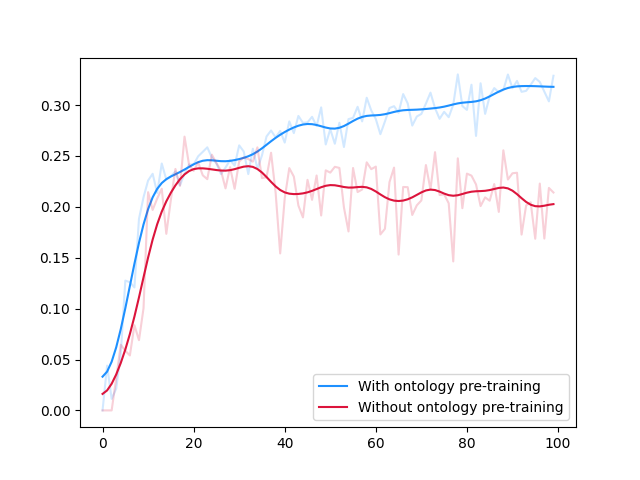}
    \caption{F1 score (micro) / original TOX21 dataset}
    \label{fig:f1-challenge}
    \end{subfigure}
\caption{Development of ROC-AUC and F1 score (micro) during training on the validation sets of the Tox21 dataset  available as part of MoleculeNet and the original TOX21 challenge} 
\end{figure*}

The final result of our training stack are four models: with or without ontology pre-training, and fine-tuned on the original Tox21 competition dataset or on the smaller version of the Tox21 dataset published as part of MoleculeNet.
The semantic pre-training already showed a clear impact during the training phase. Figures~\ref{fig:ar-moleculenet}-\ref{fig:f1-challenge} depict the curves for two metrics (F1 score and ROC-AUC) on our validation set as evaluated at the end of each epoch during training. It can be seen that models with ontology pre-training start with a better initial performance and also retain this margin throughout the training. This behaviour is also reflected in the predictive performance on both test sets. Table~\ref{tbl:tox21pred} shows the predictive behaviour for the dataset from MoleculeNet and the original challenge. The leading model (highlighted in bold) is predominantly the one that received additional ontology pre-training. This is particularly visible for the more noisy and sparse dataset used in the original Tox21 competition. The overall improved performance shows that pre-training with a more general ontology pre-training does support the network for the more specialised toxicity prediction.
The drastic drop that can be seen around epoch 50 in Figure~\ref{fig:f1-challenge} but not for the pre-trained model in Figure~\ref{fig:f1-moleculenet} further indicates that ontology pre-training hedges the model against early overfitting. The reported results, and in particular the F1 scores, however, show that there is still a large margin of error for this task.



\begin{table*}
\centering
    \begin{tabular}{l||l|l|l|l|l|l|l|l|l|l}
Dataset & \multicolumn{5}{|c|}{Tox 21 (MoleculeNet)} & \multicolumn{4}{|c|}{Tox21 (Challenge)}    \\ \hline
 Metric & \multicolumn{2}{|c|}{F1} & \multicolumn{3}{|c|}{ROC-AUC} &  \multicolumn{2}{|c|}{F1} & \multicolumn{2}{|c|}{ROC-AUC}\\  \hline
Model & \multicolumn{4}{|c|}{Our Model} & SSL-GCN &  \multicolumn{4}{|c|}{Our Model}    \\  \hline
 Ontology Pre-training & yes & no & yes & no & - & yes & no & yes & no\\  \hline  \hline
NR-AR & 0.41 & \textbf{0.52} & \textbf{0.82} & 0.76 & 0.80 & 0.1 & \textbf{0.14} & \textbf{0.63} & 0.62\\
NR-AR-LBD & \textbf{0.51} & 0.5 & \textbf{0.85} & 0.77 &  0.76 & 0.05 & \textbf{0.1} & \textbf{0.69} & 0.67\\
NR-AhR & \textbf{0.53} & 0.45 & 0.81 & 0.82  & \textbf{0.83} & \textbf{0.23} & 0.05 & \textbf{0.8} & 0.69\\
NR-Aromatase & \textbf{0.33} & 0.15 & \textbf{0.84} & 0.8 & 0.73 & \textbf{0.25} & 0.04 & \textbf{0.75} & 0.69\\
NR-ER & \textbf{0.44} & 0.4 & \textbf{0.74} & 0.71 & 0.72  & \textbf{0.16} & 0.09 & \textbf{0.64} & 0.62\\
NR-ER-LBD & \textbf{0.37} & 0.3 & \textbf{0.84} & 0.76 & 0.69 & \textbf{0.14} & 0.12 & \textbf{0.66} & 0.63\\
NR-PPAR-gamma & \textbf{0.29} & - & \textbf{0.84} & 0.83 & 0.76 & \textbf{0.14} & - & \textbf{0.67} & 0.66\\
SR-ARE & 0.48 & \textbf{0.53} & 0.8 & \textbf{0.84} & 0.73 & \textbf{0.37} & 0.23 & \textbf{0.71} & 0.69\\
SR-ATAD5 & 0.14 & \textbf{0.19} & \textbf{0.75} & 0.74 & 0.72 & \textbf{0.16} & - & \textbf{0.65} & 0.65\\
SR-HSE & \textbf{0.24} & 0.22 & 0.82 & \textbf{0.82} & 0.78 & \textbf{0.13} & 0.09 & \textbf{0.76} & 0.68\\
SR-MMP & \textbf{0.62} & 0.53 & \textbf{0.9} & 0.88 & 0.81 & \textbf{0.48} & 0.21 & \textbf{0.86} & 0.82\\
SR-p53 & \textbf{0.39} & 0.35 & \textbf{0.83} & 0.8 & 0.75 & \textbf{0.3} & - & \textbf{0.82} & 0.78\\
\end{tabular}
\caption{Class-wise scores on the test set on both Tox21 datasets. Bold values denote the best value for a particular combination of dataset and metric. NR - nuclear receptor; AR - androgen receptor; LBD - luciferase; AhR - aryl hydrocarbon receptor; ER - estrogen receptor; PPAR - peroxisome proliferator-activated receptor; SR - stress response; ARE - nuclear factor antioxidant response; ATAD5 - genotoxicity; HSE - heat shock factor response; MMP - mitochondrial response; p53 - DNA damage response. 
\label{tbl:tox21pred}}
\end{table*}


\subsection{Interpretability}
\label{ssec:interpretability}


Attention weights in Transformer networks can be visualised to enable a kind of salience-based visual interpretation for predictions directly connected with the input data \cite{vig_bertology_2021}. Previous work \cite{glauer_interpretable_2022} explored the link between attention weights  to the prediction of ontology classes, and observed that the network learned to pay attention to relevant substructures when making predictions of ontology classes to which the molecule belongs. 

In the current work, our hypothesis was that the additional information from the ontology would both enhance the prediction and enhance the coherence of the attention weights for explanatory visualisations. To test this hypothesis we explored the attention visualisations for the predictions by the ontology pre-trained network as compared to the normal network. 

\begin{figure}
    \centering
    \begin{subfigure}{0.6\columnwidth}
    \centering
    \includegraphics[width=\columnwidth]{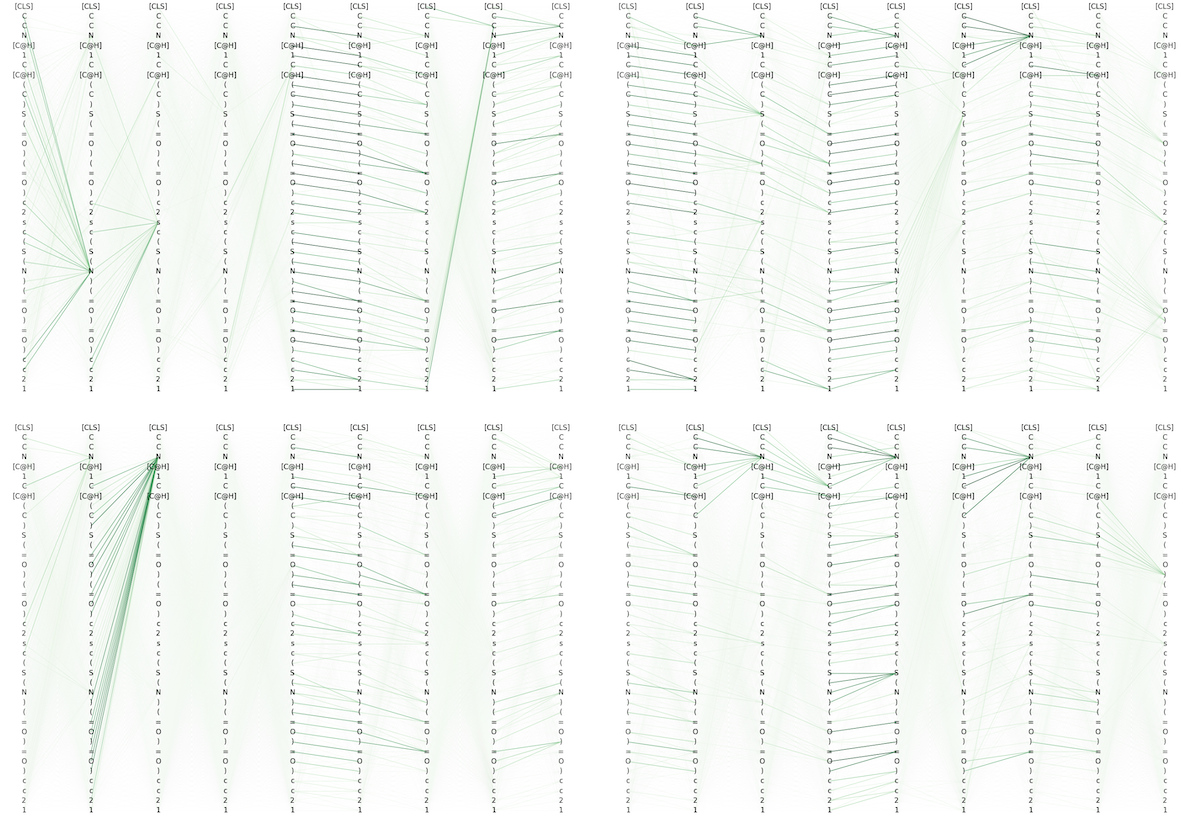} \\
    \caption{Attention plots for layers 2-3}
   \label{fig:attentionweights-123}
    
    \end{subfigure}
    \begin{subfigure}{0.6\columnwidth}
    \centering
    \includegraphics[width=\columnwidth]{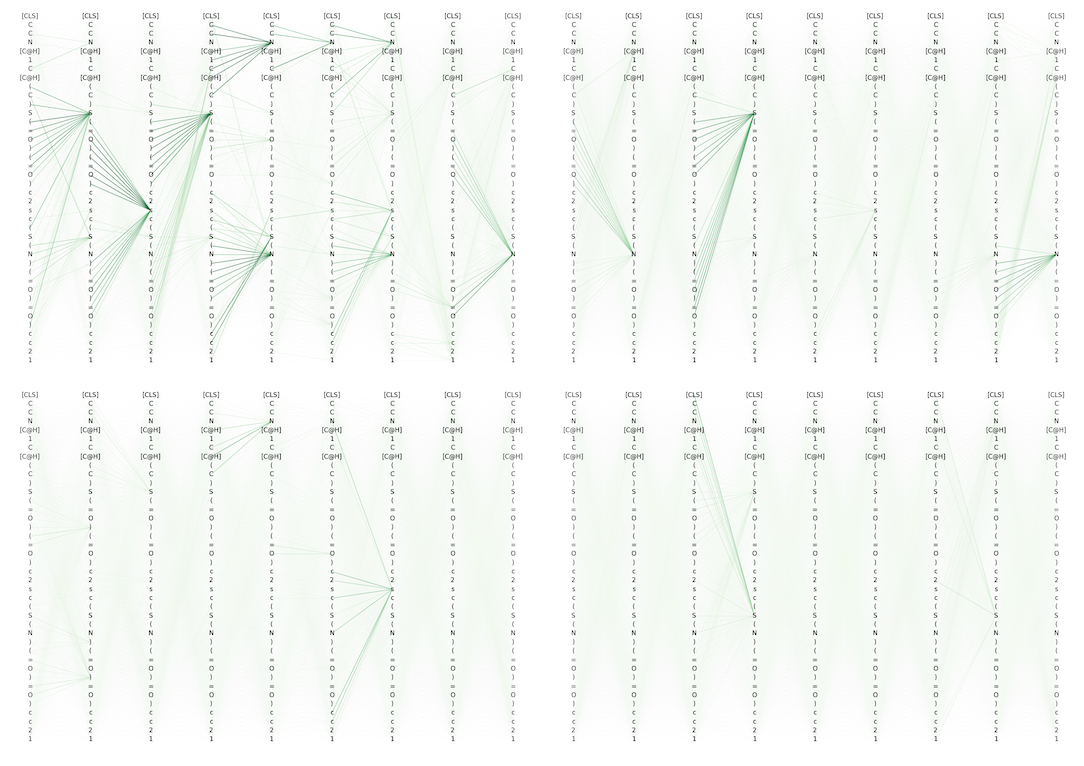} \\
    \caption{Attention plots for layers 4-5}
    \label{fig:attentionweights-456}
    \end{subfigure}
    
    \begin{subfigure}{\columnwidth}
    \centering
    \includegraphics[width=0.6\columnwidth]{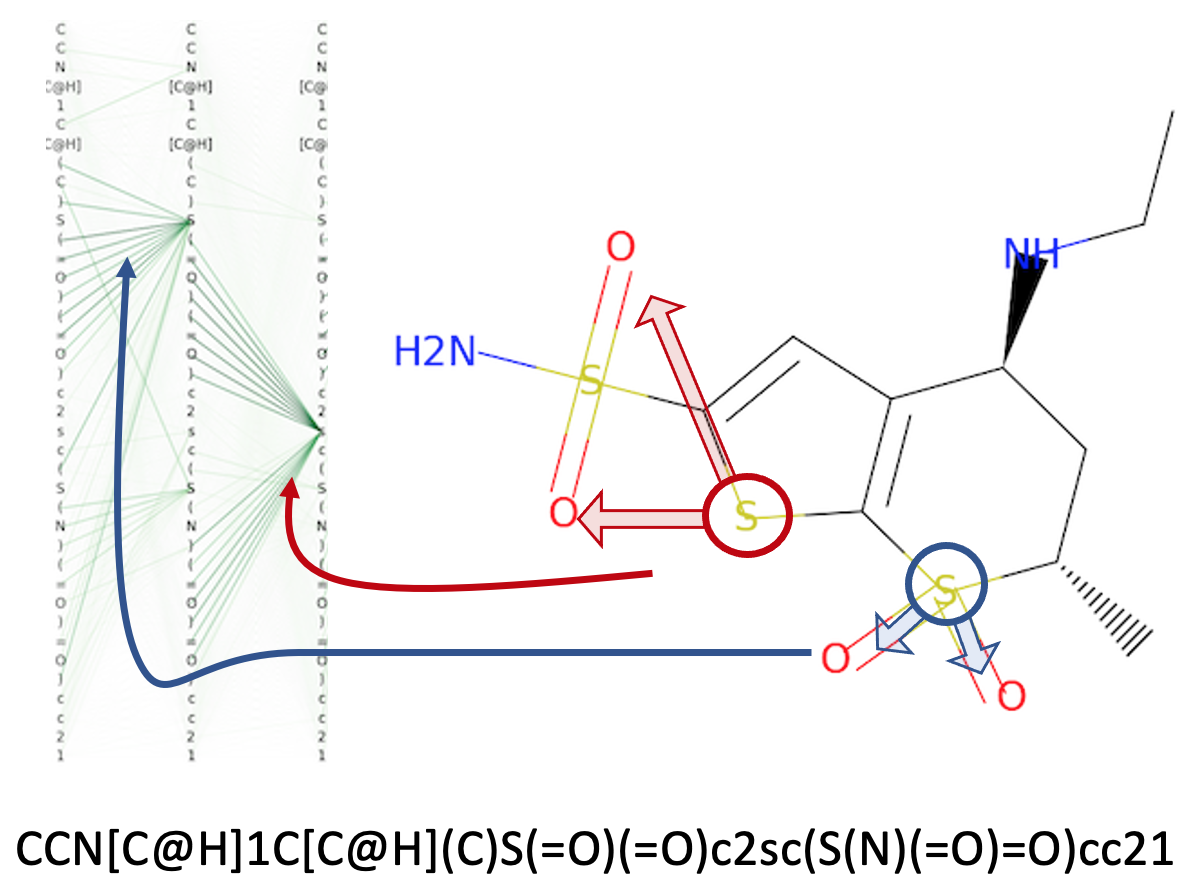}
    \caption{Attention clusters relate to the molecule structure}
    \label{fig:attentionweights-mol}
    \end{subfigure}

    \caption{Visualisation of the attention weights for the layers 2-3 in subfigure a) and layers 4-5 in subfigure b). We omit layers 1 and 6 as they had largely uniform attention weights. Each subfigure compares the ontology pre-trained network (first row) to the prediction network without pre-training (second row). c) The molecular structure processed in the attention plots is depicted with attention from layer 4 heads 1-2, showing how attention clusters relate to the molecular structure. }
    \label{fig:attentionweights}
\end{figure}

Figure~\ref{fig:attentionweights} shows an individual example of the attention weight that the network uses when visiting specific input tokens. The molecule depicted is TOX25530, corresponding to the sulfonamide anti-glaucoma drug dorzolamide, which is not toxic. Dark green lines indicate strong, focused attention, while more opaque lines indicate broader, less focused attention. As this example illustrates, we observed that the ontology pre-trained network often shows more coherence and structure in its attention weights compared to the baseline network without ontology pre-training. This is reflected in the triangular clusters of darker attention weights in the top rows of Figure~\ref{fig:attentionweights}a and b. Clusters reflect that from a particular position in the input token sequence, strong attention is placed on a subset of the molecule, reflecting relevant substructures within the molecule. Figure~\ref{fig:attentionweights}c shows how the attention weights relate to the molecular graph for this molecule. Attention weight relationships may be short-range (nearby atoms or groups) or long-range (atoms or groups further away within the molecule). 


\begin{table}
\centering
\begin{tabular}{l|p{1.5cm}|p{1.5cm}}
    & \multicolumn{2}{c}{Ontology Pre-training} \\
    & yes & no \\ \hline
    Tox21 Challenge & 0.86 & 0.90 \\
    Tox21 MoleculeNet & 0.79 & 0.85 
\end{tabular}
\caption{Entropy values averaged over all tokens, attention heads, layers and atoms in the respective test set.}
\label{tbl:entropy}
\end{table}

To test this visual intuition more systematically, we computed the entropy for each attention head in each layer. Attention is computed using softmax and can therefore be interpreted as a probability distribution. In order to evaluate our hypothesis that the ontology pre-training also impacts the way a model focuses its attention, we calculated the average entropy of these distributions. Entropy is a measure for the randomness of a distribution. A perfectly even distribution would result in a entropy value of 1 (complete uncertainty), while a distribution in which only one event can possibly occur will result in an entropy value of 0 (complete certainty). That means that an attention distribution with an entropy value of 1 is not paying attention to any particular part of the molecule, but is spreading attention evenly. An entropy value of 0 indicates that the model paid attention to only a single token of the molecule. Table~\ref{tbl:entropy} shows the aggregated entropy values for our models. It can be seen that the model that received additional ontology pre-training has a lower entropy value for both datasets. That means, that the attention is generally spent less evenly and is therefore more focused in comparison to the model that did not receive that additional training. This indicates that the ontology pre-trained model's decisions are based on more concise substructures within the molecule.


\section{Discussion}
\label{sec:discussion}

\subsection{Significance}
\label{ssec:significance}
Our approach introduces a new way to embed knowledge from an ontology into a neural network. The underlying assumption of this approach is the following: A well-designed ontology represents a classification of its domain  that has proven useful for the various goals and tasks of experts in that area. 
Thus, it is possible (and even likely) that some of the classes of the ontology reflect features that are relevant for a prediction task in that domain. 
For example, the classification of chemical entities in  ChEBI is based on empirical knowledge about the  pertinent features of molecules and their chemical behaviour, and it is reasonable to expect that some of these pertinent features are, at least indirectly, correlated with toxicity.
The goal of ontology pre-training is to enable  a model to benefit from the knowledge represented in an ontology by training it to categorise its input according to the class hierarchy from the ontology. 


This approach is applicable in any case where a dataset is available that links the input datatype for the prediction task to the classification hierarchy from the ontology. In the case of our example, the SMILES input structures are directly associated with the leaf classes of the ChEBI ontology, thus we can prepare a training dataset directly from the ontology. However, for other ontologies, the dataset may be assembled from ontology annotations which are external to the ontology but serve the purpose of linking the ontology classes to examples of instances of the corresponding input datatype.

The results in Table~\ref{tbl:tox21pred} show 
that we were able to improve the performance of our model for  toxicity 
prediction with the help of ontology pre-training. 
The inspection of the attention head weights indicates that the system indeed learned meaningful aspects of life science chemistry from the pre-training task. Further, as we will discuss next, the performance of the ontology pre-trained model compares favourably with the state of the art.

\subsection{Related work}
\label{sec:relatedWork}

\subsubsection{Toxicity prediction}
The prediction, based on chemical structure, of whether a molecule has the potential to be harmful or poisonous to living systems, is a challenging task for life science chemistry \cite{cavasotto_machine_2022,yang_silico_2018}. 
The Tox21 dataset has become a widely used benchmark for evaluating machine learning approaches to this task, thus there have been multiple previous publications using different strategies. Most approaches to toxicity prediction supplement the input data chemical structures with additional calculated features, such as known toxicophores, in order to enhance performance on the toxicity prediction task. This was the case for the winner of the original Tox21 challenge, who used a deep learning implementation together with three other types of classifier in an ensemble \cite{mayr_deeptox_2016}, and more recently \cite{peng_top_2020}, which augments the input molecular structure with physico-chemical properties. Another recent approach uses `geometric graph learning' \cite{jiang_ggl-tox_2021} which augments the input molecular graphs with multi-scale weighted coloured graph descriptors. Some approaches additionally augment the training data with more examples in order to mitigate the fact that the Tox21 dataset is small. In \cite{idakwo2020structure}, chemical descriptors were calculated and in addition, a strategy to remove class imbalance was applied including over-sampling from the minority classes in the dataset followed by cleaning of mislabelled instances. In these cases, which use a different input dataset whether through feature augmentation, data augmentation or data cleaning, we cannot directly compare their results to ours. We anticipate that feature and data augmentation approaches such as these would potentially improve our method's performance as well, but we would need to develop a more complex attention visualisation mechanism to operate across augmented inputs. Since our objective is rather to describe a new approach to incorporating ontology knowledge into a neural network, we here focus our performance evaluation on those approaches that are more directly comparable to ours. 

\cite{chenchemical2021} uses a graph neural network and a semi-supervised learning approach known as Mean Teacher which augments the training data with additional unlabelled examples. This network and training approach achieving a ROC-AUC score of 0.757 in the test set, which our approach outperforms without data augmentation. Table~\ref{tbl:tox21pred} shows a comparison of the ROC-AUC values achieved by our model against those reported for the best model (SSL-GCN) reported in \cite{chenchemical2021}. With the exception of one class, our model shows better performance for all target classes. 

ChemBERTa \cite{chithrananda2020chemberta} is the most similar to our approach in that it also uses a Transformer-based network. Its original publication also contains an evaluation on the p53 stress-response pathway activation (SR-p53) target of the Tox21 dataset from MoleculeNet. Our model exceeds the reported ROC-AUC value (ChemBERTa: 0.728, our model: 0.83).

\subsubsection{Knowledge-aware pre-training with an ontology}

Approaches to add knowledge from an ontology into a machine learning model follow several different strategies.

The most common is that the information from the ontology is used to supplement the input data in some form, such as by adding synonyms and classification parent labels to the input data. For example, in \cite{sahoo_ontology-based_2022} an ontology is used to supplement the input data with an ontology-based `feature engineering' strategy. 

A second approach is that the ontology content is itself provided as input that is embedded into the network, for example by using random walks through the ontology content to create sentences representing the structure of the ontology for input into a language model. Ontology embedding strategies include OWL2Vec$^*$ \cite{DBLP:journals/ml/ChenHJHAH21} and OPA2Vec \cite{smaili_opa2vec_2019}. These approaches are suitable for tasks such as knowledge graph completion or link prediction, but the additional information provided by such embeddings is not inherently connected in the internal representation space to the other information learned by the network, and this limits their potential benefit if the input datatype is complex and internally structured. For example, in the chemistry case, the information about the molecular structure of the chemical that is provided in the SMILES input strings would not be connected intrinsically to the information about the class hierarchy provided to the network by an ontology embedding strategy. 

There are some examples of the use of biological ontologies together with biomolecular input data types that are closer to our approach. 
OntoProtein \cite{DBLP:conf/iclr/ZhangBL0HDZLC22} combines background knowledge from the Gene Ontology with protein sequences to improve the prediction of protein functions.
OntoProtein uses the Gene Ontology in pre-training a protein embedding. Existing protein embeddings such as ProtBERT are enhanced by embedding the Gene Ontology as a knowledge graph (following approaches such as OWL2Vec$^*$) and then explicitly aligning the two embeddings. By contrast, our approach uses the ontology more directly. Namely, our pre-training exploits both ChEBI's class structure as well its class hierarchy. The class structure leads to an aggregation of molecules with similar sub-structures and properties, which can enhance the learning process. The class hierarchy is indirectly influencing the learning process as well, because a subclass relation corresponds to a subset relation for the training samples. OntoProtein uses the subclass hierarchy only for defining depth of ontology terms, which influences learning only very indirectly. Hence, our model incorporates expert knowledge in a more direct way.
In the future, we will try to incorporate OntoProtein's approach of contrastive learning using knowledge-aware negative sampling into our approach.

Other approaches have developed custom architectures for the incorporation of knowledge from an ontology into the network. For example, DeepPheno \cite{kulmanov_deeppheno_2020} predicts phenotypes from combinations of genetic mutations, where phenotypes are encoded into the network through a novel hierarchical classification layer that encodes almost 4,000 classes from the Human Phenotype Ontology together with their hierarchical dependencies as an ontology prediction layer that informs the remainder of the training of the network. A similar approach is used in \cite{zha_ontology-aware_2022}, in which an ontology layer adds information from an ontology to a neural network to enhance the prediction of microbial biomes. Instead our approach uses the generic architecture of a standard Transformer network and learns the information from the ontology through an additional pre-training task.

\subsection{Limitations}
\label{ssec:limitations}

Our current approach only uses a fraction of the information available in the ChEBI ontology, since we only consider the structural classification of chemical entities beneath the `molecular entity' class. 
Hence, we currently consider neither classes of biological and chemical roles nor pharmaceutical applications. These classes have the potential to further enhance the knowledge that is provided to the network, and will be explored in future work.  

Another limitation is related to the way we create the ontology pre-training task: We  use the leaf nodes of ChEBI as examples for training a model to predict the subsumption relation for more general classes. 
Or, to put it differently, while from an ontological perspective the leaf nodes of ChEBI are  classes in ChEBI's taxonomy, we are treating them as instances and, thus, turning  subsumption prediction  into a classification task from a machine learning perspective. Consequently, while we use the whole structural hierarchy of ChEBI for creating the pre-training task, the model learns to classify only 856 classes, those that have a sufficient number of example structures to learn from, which is a relatively small number compared to the number of classes in ChEBI.  
Further, this approach of creating a subsumption prediction pre-training task requires rich structural annotations linked to the learning input datatype (which is the SMILES in our example case), which many ontologies do not contain.  

As indicated in Section~\ref{ssec:significance}, both of these limitations may be addressed by  
using class membership prediction for ontology pre-training. 
All that is required for ontology pre-training is a dataset 
that (a) is of the same input datatype as the fine-tuning task, (b) is annotated with terms from the ontology, and (c) contains sufficient training examples to train the model to classify the input with the terms of the ontology.  
Because we treat subsumption prediction as a classification problem anyway, both approaches are functionally equivalent.  However, using an external dataset for training (instead of generating it from the ontology itself), has the benefit that the ontology pre-training might cover the whole ontology instead of just a subset of the ontology. Further, this approach does not rely on the existence of appropriate annotations in the associated input data type.  

A further limitation of the approach is that the interpretability offered by the attention weights is limited to visual inspection. In future work we aim to develop an algorithm that is able to systematically determine clusters of attention mapped to the relevant parts of the input molecular structure. 

\section{Conclusion}

This paper presents the results of training an Electra model for toxicity prediction using the Tox21 dataset as a benchmark. The model was able to achieve state-of-the-art performance on the task of toxicity prediction, outperforming comparable models that have been trained on the same dataset.

While improving the state of the art of toxicity prediction is in itself an interesting result, 
the main contribution of the paper is the presentation of a novel approach to combining symbolic and connectionist AI. This is because our
result was achieved with the help of an additional pre-training step, which trains the model with the help of background knowledge from an ontology. 
For the presented work the pre-training task consisted of predicting subclass relationships between classes in the ChEBI ontology, but other tasks (e.g., predicting class membership) are likely to be equally suitable to achieve the purpose of ontology pre-training, namely, to train the model to recognise the meaningful classification of the entities in a given domain, as represented by the ontology.  

As we have illustrated in this paper, ontology pre-training has the potential benefit of reducing the time needed for fine-tuning and improving performance. Further, as we have illustrated  in Section~\ref{ssec:interpretability}, an 
inspection of attention heads indicates that some of the attention patterns of the model  correlate to the substructures that are pertinent for chemical categories in ChEBI. 
Thus, since ontological categories are meaningful for humans, another potential benefit of ontology pre-training is an improved  interpretability of the trained model. 
In the future, we are planning to further investigate this line of research by systematically analysing the attention patterns of the pre-trained model and automatically linking these patterns to the ontology. 

As we discussed in Section~\ref{ssec:limitations}, currently we are only using some of the knowledge available in the ontology for pre-training purposes. In particular, we do not include classes from other parts of the ontology, nor do we include other axioms aside from the hierarchy. In the future we are planning to overcome these limitations by sampling a wider set of classes from ChEBI and by using a more complex architecture that combines a Transformer with a logical neural network (LNN) \cite{riegel2020logical}.
The LNN is able to represent logical axioms from the ontology as a first-order theory, translated from OWL~\cite{FOWL22}. This will enable us to use logical axioms from the ontology (and therefore also its binary relations) to influence both the ontology pre-training as well as the fine-tuning.





\bibliographystyle{plain}
\bibliography{toxpred}

\end{document}